\documentclass[runningheads]{llncs}
\usepackage{graphicx}
\usepackage{booktabs}
\usepackage{multirow}
\usepackage{amssymb}
\usepackage{amsfonts}
\usepackage{subfig}
\usepackage{enumitem}
\usepackage{xcolor}
\usepackage{comment}

\usepackage{hyperref}

\usepackage{xspace}

\usepackage{footnote}

\begin{document}

\raggedbottom

\title{Long-Tailed Classification of Thorax Diseases on Chest X-Ray: A New Benchmark Study}
\titlerunning{Long-Tailed Classification of Thorax Diseases on Chest X-Ray}
%
\author{Gregory Holste\inst{1} \and
Song Wang\inst{1} \and
Ziyu Jiang\inst{2} \and
Thomas C. Shen\inst{3} \and
George Shih\inst{4} \and
Ronald M. Summers\inst{3} \and
Yifan Peng\inst{4}\thanks{Denotes joint corresponding authorship.} \and
Zhangyang Wang \inst{1}\textsuperscript{$\star$}}
\authorrunning{G. Holste \textit{et al.}}
%
\institute{\textsuperscript{1}The University of Texas at Austin, Austin, TX, USA \\ \textsuperscript{2}Texas A\&M University, College Station, TX, USA \\
\textsuperscript{3}National Institutes of Health, Bethesda, MD, USA \\ \textsuperscript{4}Weill Cornell Medicine, New York, NY, USA \\
\email{yip4002@med.cornell.edu, atlaswang@utexas.edu}}

\maketitle              

\begin{abstract}
Imaging exams, such as chest radiography, will yield a small set of common findings and a much larger set of uncommon findings. While a trained radiologist can learn the visual presentation of rare conditions by studying a few representative examples, teaching a machine to learn from such a “long-tailed” distribution is much more difficult, as standard methods would be easily biased toward the most frequent classes. In this paper, we present a comprehensive benchmark study of the long-tailed learning problem in the specific domain of thorax diseases on chest X-rays. We focus on learning from naturally distributed chest X-ray data, optimizing classification accuracy over not only the common ``head" classes, but also the rare yet critical ``tail” classes. To accomplish this, we introduce a challenging new long-tailed chest X-ray benchmark to facilitate research on developing long-tailed learning methods for medical image classification. The benchmark consists of two chest X-ray datasets for 19- and 20-way thorax disease classification, containing classes with as many as 53,000 and as few as 7 labeled training images. We evaluate both standard and state-of-the-art long-tailed learning methods on this new benchmark, analyzing which aspects of these methods are most beneficial for long-tailed medical image classification and summarizing insights for future algorithm design. The datasets, trained models, and code are available at \url{https://github.com/VITA-Group/LongTailCXR}.

\end{abstract}

\section{Introduction}
\raggedbottom

Like most diagnostic imaging exams, chest radiography produces a few very common findings, followed by many relatively rare findings \cite{Paul2021Zeroshot,Zhou2021Review}. Such a ``long-tailed" (LT) distribution of outcomes can make it challenging to learn discriminative image features, as standard deep image classification methods will be biased toward the common ``head" classes, sacrificing predictive performance on the infrequent ``tail" classes \cite{zhang2021deep}. In other settings and modalities, there are a select few examples of LT datasets, such as in dermatology \cite{liu2020deep} and gastrointestinal imaging \cite{borgli2020hyperkvasir}; however, the data from Liu et al~\cite{liu2020deep} are not publicly available, and the \textit{HyperKvasir} dataset \cite{borgli2020hyperkvasir} -- while providing 23 unique class labels with several very rare conditions (\textless $50$ labeled examples) -- only contains about 10,000 labeled images for classification. Additionally, while many studies offer techniques to combat class imbalance for medical image analysis problems \cite{marrakchifighting2021,zhuangcare2019,linautomated2021,galdranbalanced2021}, very few methods specifically address the challenges posed by an LT distribution, as there is no freely available benchmark for this purpose.

Only recently have studies begun to use the lens of ``LT learning" to describe and improve medical image understanding solutions. For example, Galdran et al. \cite{galdranbalanced2021} proposed Balanced-MixUp, an extension of the MixUp \cite{zhang2018mixup} regularization technique with class-balanced sampling, a common approach in the LT learning literature. Ju et al. \cite{jurelational2021} grouped rare classes into subsets based on prior knowledge (location, clinical presentation) and used knowledge distillation to train a ``teacher" model to enforce the ``student" to learn these groupings. Zhang et al. \cite{zhangmbnm2021} combined a feature ``memory" module, resampling of tail classes, and a re-weighted loss function to improve the LT classification of several medical datasets. More broadly, many relevant techniques have been developed in the related fields of imbalanced learning \cite{zhuangcare2019,marrakchifighting2021} and few-shot learning \cite{quellecautomatic2020,lidifficulty2020}. These medical image-specific techniques, plus the wealth of methods from the computer vision literature \cite{chawla2002smote,huang2016learning,wang2017learning,linfocal2017,cuiclass2019,caolearning2019,shu2019meta,kangdecoupling2020,zhang2021deep,jiang2021self,park2021influence,kini2021label}, provide a foundation from which the medical deep learning community can develop methods for medical LT classification.

Since no large-scale, publicly available dataset exists for LT medical image classification, we curate a large benchmark (\textgreater 200,000 labeled images) of two thorax disease classification tasks on chest X-rays. Further, we evaluate state-of-the-art LT learning methods on this data, analyzing which components of existing methods are most applicable to the medical imaging domain.

Our contributions can be summarized as follows:
\begin{itemize}[topsep=1pt, leftmargin=1.25cm]
    \item[$\bullet$] We formally introduce the task of long-tailed classification of thorax disease on chest X-rays. The task provides a comprehensive and realistic evaluation of thorax disease classification in clinical practice settings.
    \item[$\bullet$] We curate a large-scale benchmark from the existing representative datasets NIH ChestXRay14 \cite{wang2017learning} and MIMIC-CXR \cite{johnson2019mimic}. The benchmark contains five new, more fine-grained pathologies, producing a challenging and severely imbalanced distribution of diseases. We describe the characteristics of this benchmark and will publicly release the labels.
    \item[$\bullet$] We find that the standard cross-entropy loss and augmentation methods such as MixUp fail to adequately classify the rarest ``tail" classes. We observe that class-balanced re-weighting improves performance on infrequent classes, and ``decoupling" via classifier re-training is the most effective approach for both datasets.
\end{itemize}

\begin{figure}[!ht]
    \centering
    \includegraphics[scale=0.41]{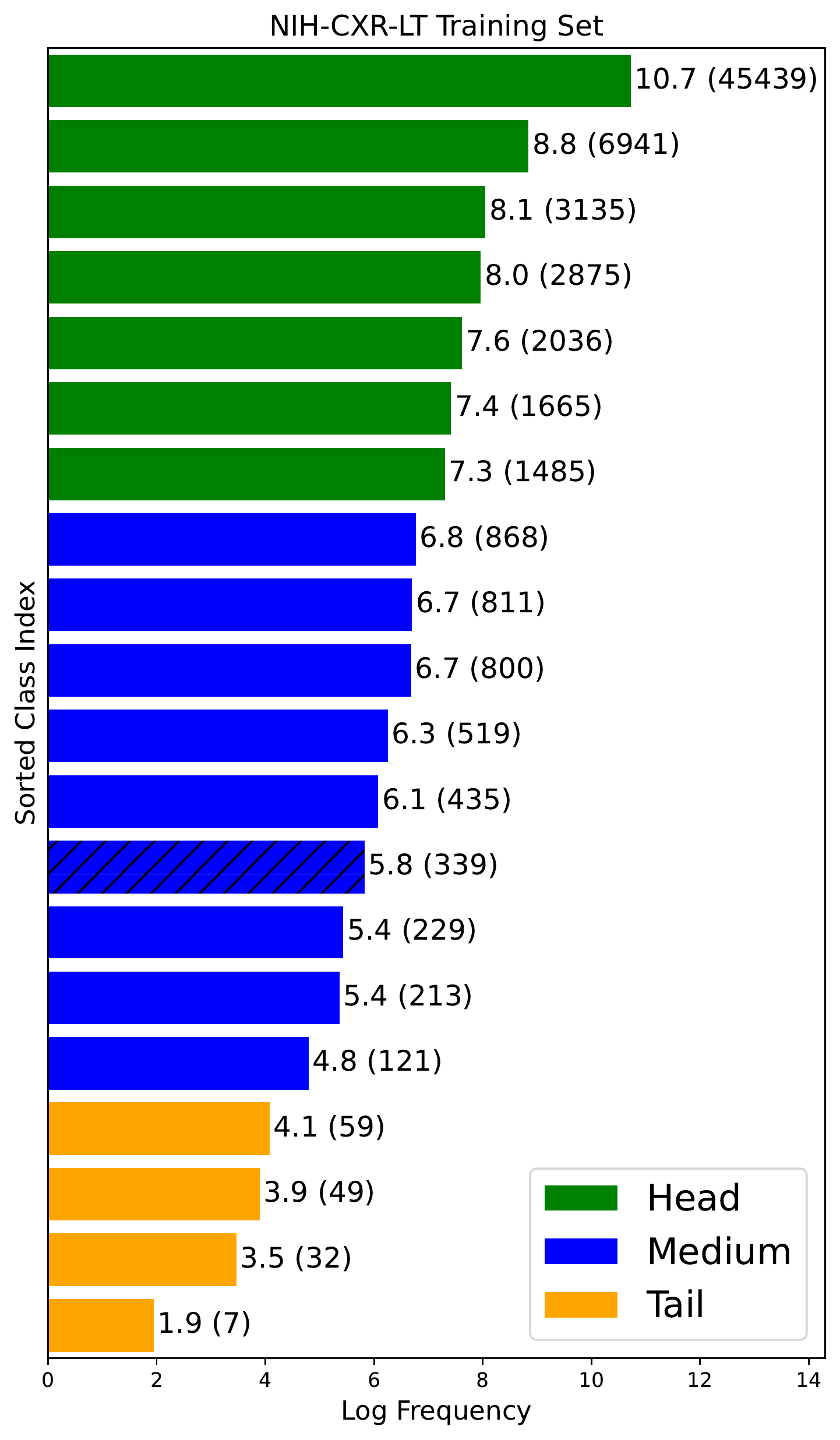}
    \includegraphics[scale=0.41]{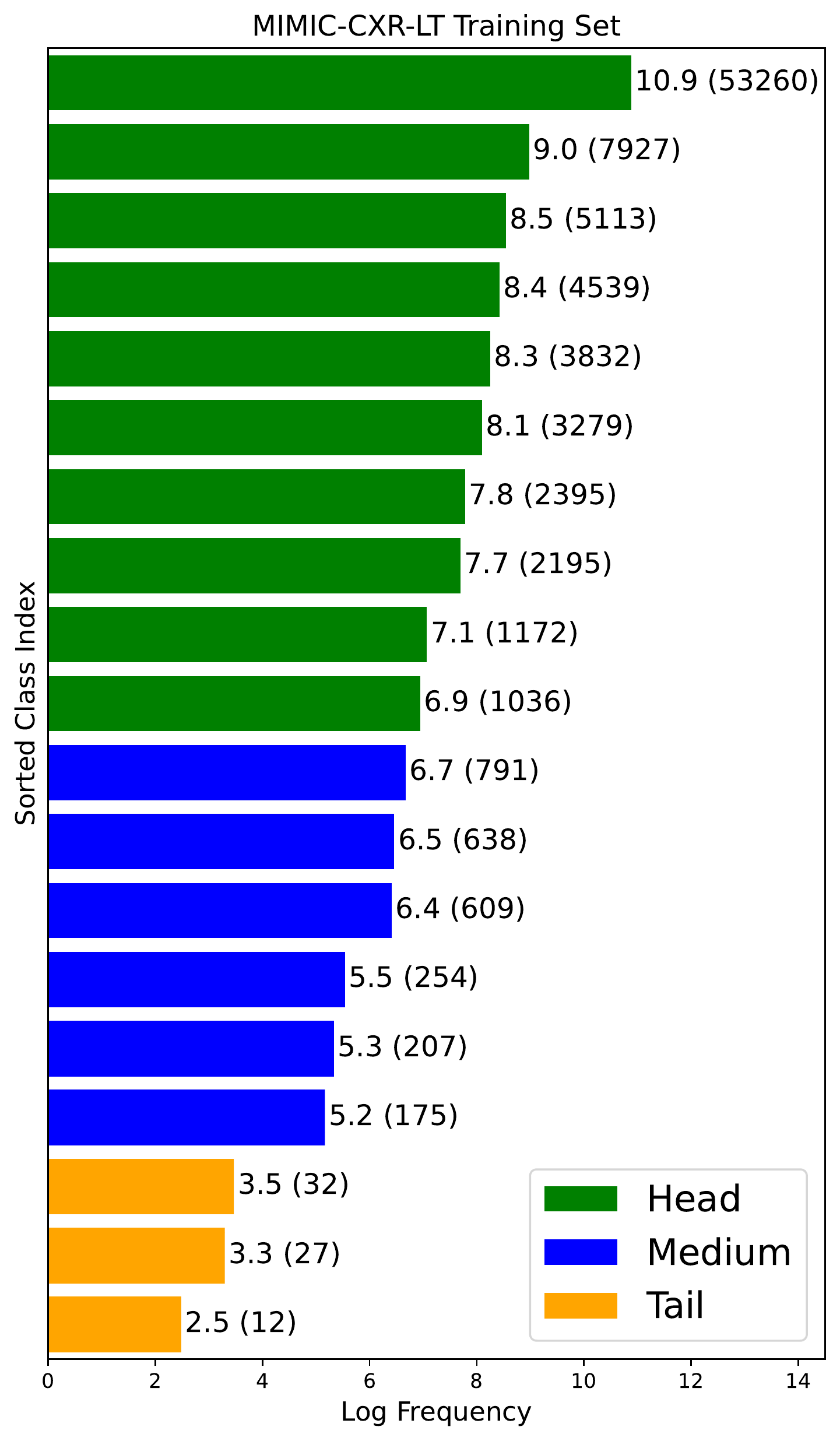}
    \caption{Long-tailed distribution of thorax disease labels for the proposed NIH-CXR-LT (left) and MIMIC-CXR-LT (right) training datasets. Values by each bar represent log-frequency, while values in parentheses represent raw frequency. Textured bars represent newly added disease labels, which help create naturally long-tailed distributions without the need for artificial subsampling.}
    \label{fig:data}
\end{figure}

\section{Long-Tailed Classification of Thorax Diseases}
\raggedbottom
\subsection{Task Definition}
Disease patterns in chest X-rays are numerous, and their incidence exhibits a long-tailed distribution \cite{Zhou2021Review,Paul2021Zeroshot}: while a small number of common diseases have sufficient observed cases for large-scale analysis, most diseases are infrequent. Conventional computer vision methods may fail to correctly identify uncommon thorax disease classes due to the extremely imbalanced class distribution \cite{Paul2021Zeroshot}, introducing a new and clinically valuable LT classification task on chest X-rays. We formulate the LT classification task first by dividing thorax disease classes into ``head" (many-shot: \textgreater 1,000), ``medium" (medium-shot: 100-1000, inclusive), and ``tail" (few-shot: \textless 100) categories according to their frequency in the training set.

\subsection{Dataset Construction}
We curate two long-tailed chest X-ray benchmarks, \textbf{NIH-CXR-LT} and \textbf{MIMIC-CXR-LT}, for NIH ChestXRay14 \cite{wangchest2017} and MIMIC-CXR \cite{johnson2019mimic}, respectively. Each study of NIH ChestXRay14 and MIMIC-CXR usually contains one or more chest radiographs and one free-text radiology report. To generate a strongly long-tailed distribution without artificially subsampling, we introduce five new rare disease findings that are text-mined from radiology reports: Calcification of the Aorta, Subcutaneous Emphysema, Tortuous Aorta, Pneumomediastinum, and Pneumoperitoneum.
We identify the presence or absence of new disease findings by parsing the text report associated with each study following the method detailed in RadText \cite{Peng2018NegBioAH,wangchest2017}.

For this study, we only use frontal-view, single-label images, as most LT methods are developed specifically for multi-class (not multi-label) classification. Following the structure of previous LT benchmark datasets in computer vision, such as ImageNet-LT \cite{liu2019openlongtailrecognition}, we split NIH-CXR-LT and MIMIC-CXR-LT into \textit{training}, \textit{validation}, \textit{test}, and \textit{balanced test} sets. Since both datasets contain patients with multiple images, we split them at the patient level to prevent data leakage. Both validation and balanced test sets are small but perfectly balanced, where the balanced test set is a subset of the larger, imbalanced test set. This data split allows for evaluation consistent with the LT literature (via the balanced test set), as well as more traditional evaluation on a large naturally distributed set (via the test set). The resulting splits produce extreme class imbalance, with an \textit{imbalance factor} -- the cardinality of the most frequent training class divided by the cardinality of the least frequent training class -- of 6,491 for NIH-CXR-LT and 4,438 for MIMIC-CXR-LT. Full detailed statistics and data split for NIH-CXR-LT and MIMIC-CXR-LT can be found in the Supplementary Materials.

\textbf{NIH-CXR-LT.} NIH ChestXRay14 contains over 100,000 chest X-rays labeled with 14 pathologies, plus a ``No Findings" class. We construct a single-label, long-tailed version of the NIH ChestXRay14 dataset by introducing five new disease findings described above. The resulting NIH-CXR-LT dataset has 20 classes, including 7 head classes, 10 medium classes, and 3 tail classes. NIH-CXR-LT contains 88,637 images labeled with one of 19 thorax diseases, with 68,058 training and 20,279 test  images. The validation and balanced test sets contain 15 and 30 images per class, respectively.

\textbf{MIMIC-CXR-LT.} We construct a single-label, long-tailed version of MIMIC-CXR in a similar manner. MIMIC-CXR is a multi-label classification dataset with over 200,000 chest X-rays labeled with 13 pathologies and a ``No Findings" class. The resulting MIMIC-CXR-LT dataset contains 19 classes, of which 10 are head classes, 6 are medium classes, and 3 are tail classes. MIMIC-CXR-LT contains 111,792 images labeled with one of 18 diseases, with 87,493 training images and 23,550 test set images. The validation and balanced test sets contain 15 and 30 images per class, respectively.

\begin{table}[!ht]
    \centering
    \renewcommand{\arraystretch}{1.1}
    \caption{Long-tailed learning methods selected for benchmarking grouped by type of approach (``R" = Re-balancing, ``A" = Augmentation, ``O" = Other). ``RW" = re-weighted with scikit-learn weights \cite{scikit-learn}, ``CB" = re-weighted with class-balanced weights \cite{cuiclass2019}.}
    \begin{tabular}{lccc|lccc}
        \toprule
        Method & R & A & O & \multicolumn{1}{c}{Method} & R & A & O \\
        \midrule
        Softmax (Baseline) & & & & CB LDAM-DRW \cite{caolearning2019} & \checkmark & &\\
        
        CB Softmax & \checkmark & & & RW LDAM \cite{caolearning2019} & \checkmark & & \\
        
        RW Softmax & \checkmark & & & RW LDAM-DRW \cite{caolearning2019} & \checkmark & &\\
        
        Focal Loss \cite{linfocal2017} & \checkmark & & & MixUp \cite{zhang2018mixup} & & \checkmark &\\
        
        CB Focal Loss \cite{linfocal2017} & \checkmark & & & Balanced-MixUp \cite{galdranbalanced2021} & \checkmark & \checkmark & \\
        
        RW Focal Loss \cite{linfocal2017} & \checkmark & & & Decoupling--cRT \cite{kangdecoupling2020} & \checkmark & & \checkmark\\
        
        LDAM \cite{caolearning2019} & \checkmark & & & Decoupling--$\tau$-norm \cite{kangdecoupling2020} & \checkmark & & \checkmark\\
        
        CB LDAM \cite{caolearning2019} & \checkmark & & \\
        \bottomrule
    \end{tabular}
    \label{method:summary}
\end{table}

\subsection{Methods for Benchmarking}
In their survey, Zhang \textit{et al.} group LT learning methods into three main categories: class re-balancing, information augmentation, and module improvement \cite{zhang2021deep}. We simplify this categorization down to \underline{re-balancing}, \underline{augmentation}, and \underline{others}, noting that some sophisticated methods can fall into more than one of these categories. We have summarized our selected methods for benchmarking with their corresponding categorizations in Table \ref{method:summary}.

Class re-balancing, arguably the most common approach to LT learning, usually involves \textit{resampling} the data such that it is effectively balanced during training or \textit{re-weighting} a loss function to modulate the importance of classes based on their frequency. Resampling methods include SMOTE \cite{chawla2002smote}, which undersamples common classes and oversamples rare classes, and progressively-balanced sampling \cite{kangdecoupling2020}, which interpolates from instance- to class-balanced sampling; recent re-weighting strategies include Focal Loss \cite{linfocal2017}, Label-Distribution-Aware Margin (LDAM) Loss \cite{caolearning2019}, and Influence-Balanced Loss \cite{park2021influence}. In addition to the baseline softmax cross-entropy loss function, we consider Focal Loss and LDAM, with optional deferred re-weighting (DRW). For re-weighting strategies, we select the ``class-balanced" (CB) approach outlined in \cite{cuiclass2019} and the re-weighting approach implemented by the scikit-learn library \cite{scikit-learn}.

Approaches to ``information augmentation" can include customized data augmentation, as well as transfer learning from related data domains. For this category, we choose MixUp \cite{zhang2018mixup} and Balanced-MixUp \cite{galdranbalanced2021}. MixUp is an augmentation technique that linearly mixes pairs of input images and labels according to a Beta distribution, producing a strong regularizing effect. Balanced-MixUp, as explained earlier, is an extension of MixUp that linearly mixes pairs of images and labels, where one image is drawn from a batch of instance-balanced (naturally distributed) data and the other from class-balanced (resampled) data.

Lastly, other popular approaches to LT learning include ensembling, representation learning, classifier design, and decoupled training. For this category, we proceed with two straightforward decoupling methods: classifier re-training (cRT) and $\tau$-normalization. Kang et al. \cite{kangdecoupling2020} observed that they could achieve state-of-the-art results on several LT learning benchmarks by (1) learning representations from naturally distributed data, then (2) re-training or otherwise calibrating the classification head in order to better discriminate tail classes. After training a model on instance-balanced data, cRT freezes this trained backbone, then re-initializes and re-trains the classifier with class-balanced resampling. Directly using the model learned in step (1), $\tau$-normalization scales each classifier's learned weights by their magnitude raised to the power $\tau$.

\subsection{Experiments and Evaluation}
We evaluate the list of methods shown in Table \ref{method:summary} on NIH-CXR-LT and MIMIC-CXR-LT. To enable a fair comparison among all methods, we keep the entire training pipeline identical except for the method being applied. Specifically, we train a ResNet50 \cite{he2016deep} pretrained on ImageNet \cite{deng2009imagenet}, using the Adam optimizer 
with a learning rate of $1 \times 10^{-4}$. All models were trained for a maximum of 60 epochs with early-stopping based on overall validation accuracy. For full implementation details, refer to the Supplemental Materials and our code repository: \url{https://github.com/VITA-Group/LongTailCXR}.

We present results on both the balanced test set and imbalanced test set for each model and dataset. For the balanced test set, we report head, medium, and tail class accuracy. We additionally include the class-wise average (``overall") accuracy and the group-wise average (``avg") accuracy -- namely, the mean of the head, medium, and tail accuracy; we use this metric since we seek a model that performs well across head, medium, and tail classes regardless of how many samples or classes belong to each group. For the imbalanced test set, we report the Macro-F1 score (the unweighted mean of class-wise F1 scores) and the balanced accuracy (the accuracy with samples weighted by inverse class frequency). We choose balanced accuracy since it is resistant to class imbalance, thus necessary since the test set follows the highly imbalanced real-world data distribution.

\begin{table}[!hb]
    \centering
     \vspace{-1em}
	\renewcommand{\arraystretch}{1.1}
    \caption{Results on NIH-CXR-LT. Accuracy is reported for the balanced test set ($N=600$), where ``Avg" accuracy is the mean of the head, medium, and tail accuracy. Macro-F1 score (mF1) and balanced accuracy (bAcc) are used to evaluate performance on the imbalanced test set ($N=20,279$). The best and second-best results for a given metric are, respectively, bolded and underlined.}
    \begin{tabular}{@{}lcccccccc@{}}
        \toprule
        \multicolumn{1}{c}{Method} & \multicolumn{5}{c}{Balanced Test Set} & \multicolumn{2}{c}{Test Set} \\
        \cmidrule(lr){2-6} \cmidrule(lr){7-8}
         & Overall & Head & Medium & Tail & Avg & mF1 & bAcc \\
        \midrule
        Softmax & 0.175 & 0.419 & 0.056 & 0.017 & 0.164 & 0.131 & 0.115 \\
        CB Softmax & 0.333 & 0.295 & 0.415 & 0.217 & 0.309 & 0.177 & 0.269 \\
        RW Softmax & 0.300 & 0.248 & 0.359 & 0.258 & 0.288 & 0.116 & 0.26 \\
        Focal Loss & 0.160 & 0.362 & 0.056 & 0.042 & 0.153 & 0.142 & 0.122 \\
        CB Focal Loss & 0.303 & 0.371 & 0.333 & 0.117 & 0.274 & 0.157 & 0.232 \\
        RW Focal Loss & 0.255 & 0.286 & 0.293 & 0.117 & 0.232 & 0.090 & 0.197 \\
        LDAM & 0.232 & 0.410 & 0.133 & 0.142 & 0.228 & 0.173 & 0.178 \\
        CB LDAM & 0.295 & 0.357 & 0.285 & 0.208 & 0.284 & 0.161 & 0.235 \\
        CB LDAM-DRW & 0.377 & 0.476 & 0.356 & 0.250 & 0.361 & 0.172 & 0.281 \\
        RW LDAM & 0.353 & 0.305 & 0.419 & 0.292 & 0.338 & 0.111 & 0.279 \\
        RW LDAM-DRW & 0.370 & 0.410 & 0.367 & 0.308 & \underline{0.362} & 0.127 & \underline{0.289} \\
        MixUp & 0.170 & 0.419 & 0.044 & 0.017 & 0.160 & 0.132 & 0.118 \\
        Balanced-MixUp & 0.213 & 0.443 & 0.081 & 0.108 & 0.211 & 0.167 & 0.155 \\
        Decoupling--cRT & 0.380 & 0.433 & 0.374 & 0.300 & \textbf{0.369} & 0.138 & \textbf{0.294} \\
        Decoupling--$\tau$-norm & 0.280 & 0.457 & 0.230 & 0.083 & 0.257 & 0.144 & 0.214 \\
\bottomrule
    \end{tabular}
    \label{results:nih}
\end{table}

\begin{table}[!ht]
    \centering
	\renewcommand{\arraystretch}{1.1}
    \caption{Results on MIMIC-CXR-LT. Accuracy is reported for the balanced test set ($N=570$), where ``Avg" accuracy is the mean of head, medium, and tail accuracy. Macro-F1 score (mF1) and balanced accuracy (bAcc) are used to evaluate performance on the imbalanced test set ($N=23,550$). The best and second-best results for a given metric are, respectively, bolded and underlined.}
    \begin{tabular}{@{}lcccccccc@{}}
        \toprule
        \multicolumn{1}{c}{Method} & \multicolumn{5}{c}{Balanced Test Set} & \multicolumn{2}{c}{Test Set} \\
        \cmidrule(lr){2-6} \cmidrule(lr){7-8}
         & Overall & Head & Medium & Tail & Avg & mF1 & bAcc \\
        \midrule
            Softmax & 0.281 & 0.503 & 0.039 & 0.022 & 0.188 & 0.183 & 0.169 \\
            CB Softmax & 0.347 & 0.493 & 0.167 & 0.222 & 0.294 & 0.186 & 0.227 \\
            RW Softmax & 0.314 & 0.473 & 0.139 & 0.133 & 0.249 & 0.163 & 0.211 \\
            Focal Loss & 0.268 & 0.477 & 0.044 & 0.022 & 0.181 & 0.182 & 0.172 \\
            CB Focal Loss & 0.288 & 0.373 & 0.117 & 0.344 & 0.278 & 0.136 & 0.191 \\
            RW Focal Loss & 0.335 & 0.403 & 0.283 & 0.211 & 0.299 & 0.144 & 0.239 \\
            LDAM & 0.261 & 0.497 & 0.000   & 0.000   & 0.166 & 0.172 & 0.165 \\
            CB LDAM & 0.330  & 0.467 & 0.161 & 0.211 & 0.280  & 0.161 & 0.225 \\
            CB LDAM-DRW & 0.379 & 0.520  & 0.156 & 0.356 & \underline{0.344} & 0.197 & 0.267 \\
            RW LDAM & 0.335 & 0.437 & 0.250  & 0.167 & 0.284 & 0.149 & 0.243 \\
            RW LDAM-DRW & 0.365 & 0.447 & 0.256 & 0.311 & 0.338 & 0.177 & \underline{0.275} \\
            MixUp & 0.291 & 0.543 & 0.011 & 0.011 & 0.189 & 0.182 & 0.176 \\
            Balanced-MixUp & 0.267 & 0.480  & 0.039 & 0.011 & 0.177 & 0.176 & 0.168 \\
            Decoupling--cRT & 0.412 & 0.490  & 0.306 & 0.367 & \textbf{0.387} & 0.170  & \textbf{0.296} \\
            Decoupling--$\tau$-norm & 0.337 & 0.520  & 0.167 & 0.067 & 0.251 & 0.178 &         0.230  \\
        \bottomrule
    \end{tabular}
    \label{results:mimic}
\end{table}

\section{Results and Analysis}
For the NIH-CXR-LT dataset, the baseline method fails to adequately classify tail classes, achieving 1.7\% accuracy on those three rarest diseases (Table \ref{results:nih}). The baseline of softmax cross-entropy loss achieves a group-wise average accuracy of 0.164, but improves to 0.309 and 0.288, respectively, when using class-balanced and scikit-learn weights. Furthermore, we see that re-weighting constantly improves performance, though it is inconsistent which re-weighting method provides more significant gains than others. We also see that DRW can additionally improve performance, as evidenced by the fact that both CB LDAM-DRW and RW LDAM-DRW outperform their counterparts without DRW. We find that cRT decoupling achieves the best performance on both the balanced and imbalanced test sets, reaching 0.369 group-wise average accuracy on the balanced test set and 0.294 balanced accuracy on the test set. Classifier re-training is followed closely by RW LDAM-DRW, reaching 0.362 group-wise average accuracy and 0.294 balanced accuracy.

On MIMIC-CXR-LT, again, the baseline approach almost entirely fails to capture the tail classes, reaching 0.022 tail accuracy and 0.188 group-wise average accuracy (Table \ref{results:mimic}). Like with the NIH-CXR-LT results, re-weighting is always beneficial; for example, class-balanced re-weighting and scikit-learn re-weighting, respectively, improve focal loss performance from 0.181 to 0.278 and 0.299 group-wise average accuracy. Similarly, DRW brings even further gains to a re-weighted LDAM loss, improving group-wise accuracy by at least 0.05. Classifier re-training again achieves both the highest group-wise average accuracy on the balanced test set and the highest balanced accuracy on the test set by a considerable margin. For both the balanced and imbalanced test sets, the second-best method is a re-weighted LDAM loss with deferred re-weighting -- CB LDAM-DRW for the balanced test set and RW LDAM-DRW for the test set.

\textbf{Summary of Findings.} Overall, we see that the standard approach of optimizing softmax cross-entropy with instance-balanced weights fails to adequately capture medium and tail classes for both NIH-CXR-LT and MIMIC-CXR-LT. In contrast to the empirical success of MixUp on many natural image-based problems and Balanced-MixUp on certain medical imaging tasks, we find that MixUp and Balanced-MixUp perform similarly to the baseline for these two tasks; perhaps linearly mixing radiographs destroys valuable high-contrast signal that is necessary for discriminating disease conditions. We see that re-weighting is always beneficial, though which re-weighting method provides larger gains appears to depend on its interaction with the loss function used. We also observe that DRW can provide additional gains to standard re-weighting when used with the LDAM loss. Finally, we see that cRT decoupling was the highest-performing method on both datasets, demonstrating that decoupled training can be a simple and powerful technique for long-tailed disease classification on chest X-rays.
We note that performance is lower than prior work on the original NIH ChestXRay14 and MIMIC-CXR datasets since (1) we only consider single-label images, and (2) the newly added classes are difficult to classify and introduce confusion with the set of original diseases in each dataset.

\section{Discussion and Conclusion}
In summary, we have conducted the first comprehensive study of long-tailed learning methods for disease classification from chest X-rays. We publicly release all code, models, and data to encourage the development of long-tailed learning methods for medical image classification. While we adopted the standard practice of using ImageNet pretrained weights, this limited the list of candidate long-tailed learning methods we could use. For example, certain LT methods that use specialized architectures \cite{shu2019meta,wang2020long} or explore self-supervised learning \cite{jiang2021self,marrakchifighting2021} on other datasets are not compatible with ImageNet pretraining.
Future work will explore various pretraining options, combating long-tailed data with a different weight initialization. Lastly, future work will also involve adapting multi-label long-tailed learning methods to these datasets, acknowledging the clinical reality that patients often present with multiple pathologies at once.

\section{Acknowledgments}

This material is based upon work supported by the Intramural Research Programs of the National Institutes of Health Clinical Center, National Library of Medicine under Award No. 4R00LM013001, and National Science Foundation under Grant No. 2145640.

%
%
%
\bibliographystyle{splncs04}
\bibliography{ref}

\end{document}